\documentclass[12pt]{article}
\usepackage{graphicx}
\usepackage{amsmath}
\usepackage{hyperref}
\usepackage{booktabs}
\usepackage{float}

\usepackage{geometry}
\usepackage{scalefnt}
\usepackage[utf8]{inputenc}
\usepackage[authoryear]{natbib}

\usepackage{listings} 
\usepackage{mdframed} 
\usepackage{xcolor}   

\definecolor{promptbackground}{RGB}{240,240,240}
\definecolor{promptframe}{RGB}{200,200,200}

\lstdefinestyle{mystyle}{
    backgroundcolor=\color{promptbackground},
    frame=lr, 
    framesep=8pt, 
    framerule=0pt, 
    rulecolor=\color{promptframe}, 
    xleftmargin=10pt, 
    xrightmargin=10pt, 
    basicstyle=\ttfamily\footnotesize,
    breaklines=true,
    showstringspaces=false,
}

\newcommand{\customcite}[1]{{\itshape\color{blue}\cite{#1}}}

\begin{document}

\begin{center}
    \includegraphics[width=0.2\textwidth]{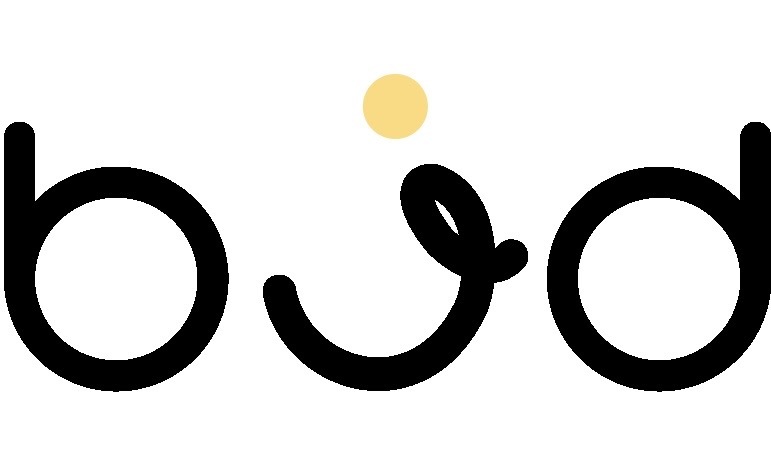} 
    \vspace{0.5cm} 
    \hrule
    \vspace{0.5cm} 
    {\Large \textbf{Intellecta Cognitiva: A Comprehensive Dataset for Advancing Academic Knowledge and Machine Reasoning}}\\[0.5cm]
    
    \hrule
    \vspace{0.5cm} 
    
    {Jithin V G, Ditto P S, Ajmal P S}\\
    {Bud Ecosystem Inc}\\[1cm]
\end{center}

\begingroup
\centering
\section*{Abstract}
\endgroup
Intellecta dataset emerges as an innovative synthetic dataset, engineered to enhance the cognitive processing capabilities of contemporary language models. With a composition of 11.53 billion tokens, integrating 8.01 billion tokens of synthetic data with 3.52 billion tokens of rich textbook data, Intellecta is crafted to foster advanced reasoning and comprehensive educational narrative generation. Leveraging the Mixtral-8x7B-Instruct-v0.1 model, the dataset facilitates the generation of complex thought processes and detailed, textbook-style explanations, thus enabling language models to engage in both critical thinking and profound educational discourse. This hybrid dataset stands as a testament to the potential of synthetic data in pushing the boundaries of AI, offering a repository that is not only vast and varied but also refined to align with ethical standards and intellectual rigor.

\bigskip 
\begin{center}
    \footnotesize 
    Dataset available at \href{https://huggingface.co/datasets/budecosystem/intellecta}{Hugging Face Intellecta Dataset}
\end{center}

\section{Design Goals}
The cornerstone of the Intellecta dataset is its commitment to enhancing the capabilities of LLMs through strategic synthetic data generation \customcite{gunasekar2023textbooks}. The primary goal is to construct a dataset with a high degree of diversity to prevent model overfitting and encourage robust generalization by following how a human learn a topic based on textbook. To actualize this, Intellecta employs targeted prompt engineering to cultivate a rich variety of data points, spanning intricate coding problems to nuanced literary analyses

Specifically, we've integrated a dynamic prompt generation system that challenges the model to explore a multitude of scenarios, preventing the homogeneity that hampers model versatility. Our data encompasses a range of complexity levels, from beginner to advanced, ensuring that models trained on this dataset are prepared for a wide spectrum of real-world applications.

The dataset's scaffolding is built on transparent, reproducible processes that adhere to open-source principles, making it a resource for collective advancement in AI. Ethical data curation is paramount, with proactive measures in place to minimize bias and ensure the dataset reflects a balanced representation of information \customcite{lee2023beyond}. These practical measures, from the dataset’s inception to its refinement, position Intellecta as an innovative and exemplary model in the field of AI data generation.

\section{Introduction}
Intellecta arises as a strategic innovation in the realm of linguistic data resources, targeting the deficit in datasets that effectively sharpen the reasoning prowess of language models. Conventional datasets often fall short in nurturing the complex cognitive functions \customcite{rane2023enhancing} that advanced language processing demands. To bridge this chasm, Intellecta offers an expansive and multifaceted dataset, curated to foster advanced reasoning capabilities in language models. Mirroring the success of the most accomplished large language models (LLMs) which thrive on synthetic data \customcite{li2022pretrained},Intellecta provides a breadth of high-quality, synthetic datasets previously inaccessible to the wider research community.

With an eye on the triumphs of prominent LLMs, Intellecta aims for extensive model generalization, aspiring to surpass existing synthetic data paradigms by controlling the entirety of the data generation process. This initiative yields a reservoir of millions of diverse data samples that span across multiple domains, thus broadening the horizon for LLMs and strengthening their capacity for diverse problem- solving. Integral to Intellecta is the incorporation of advanced synthetic generation techniques, which fabricate a dual-composed content: one that simulates complex thought processes and another that yields textbook-style elucidations laden with core concepts and pragmatic examples. We took available open source instruction data as the seed data to generate the synthetic data. From the instructions, we first create a textbook style \customcite{gunasekar2023textbooks} text explaining the concept required for answering the instruction.  The second step is to enrich the response from the seed data with thought process how the model arrive at that result. Combining these 2 will provide a textbook style concept explanation followed by exercises and the thought process to resolve the same. This correlate to the same a student learn a chapter starting with textbook followed by exercises and thought process behind it.  This synthesis forges a dataset that not only offers educational profundity but also incisive analytical clarity, catalyzing a leap forward in the domain of language modeling.

\section{Source and Curation}

\subsection{Source of Dataset}
\begin{figure}
\centering
\includegraphics[width=1\textwidth]{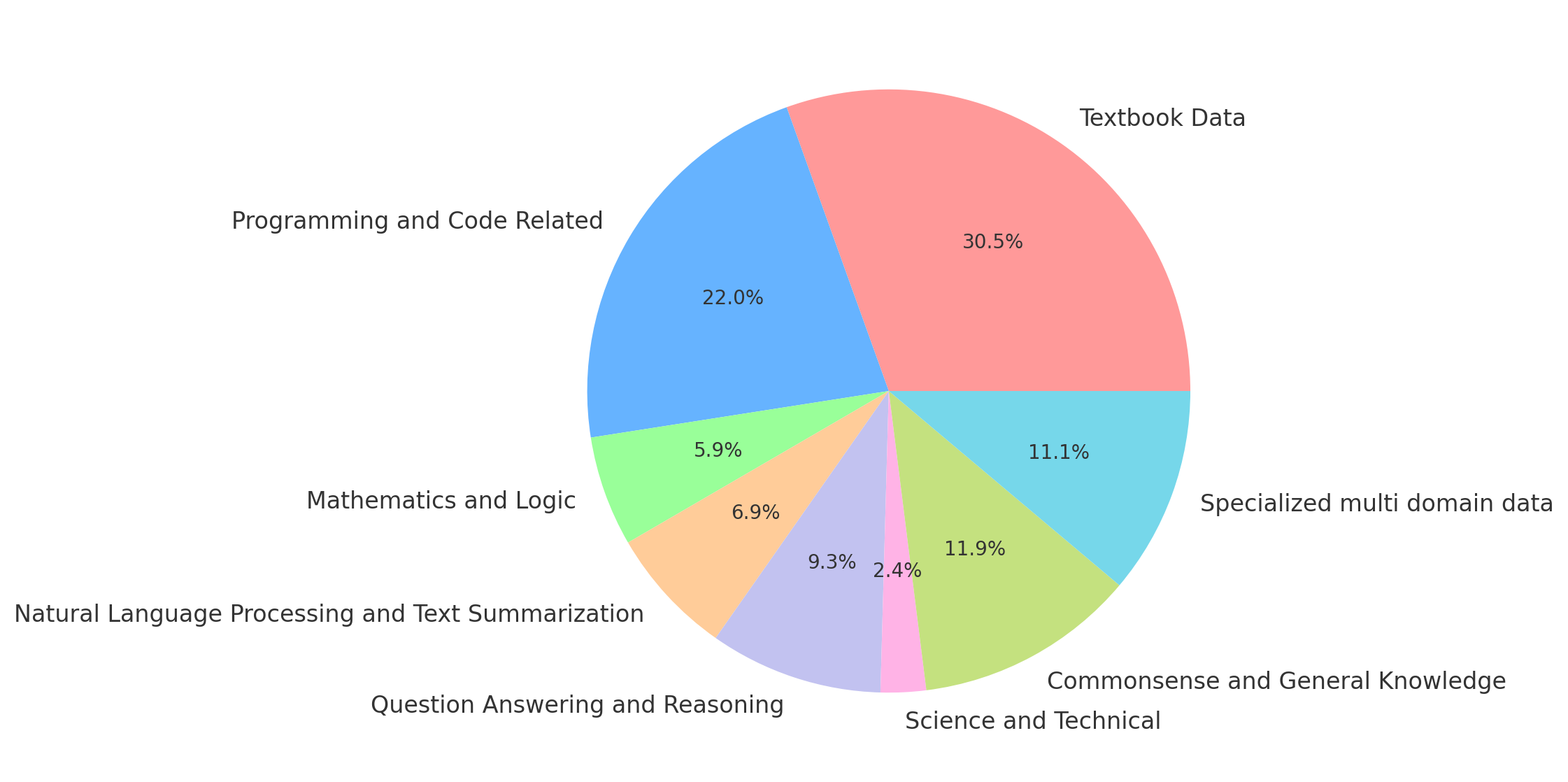}
\caption{ Distribution of Textbook and Synthetic Data in the Intellecta Dataset, highlighting the proportions of various domains within the synthetic subset, and emphasizing the dataset's comprehensive coverage across multiple disciplines.}
\label{fig:1}
\end{figure}

The Intellecta dataset presents a meticulously curated 11.53 billion token compendium, which is distinctly composed of 30.5\% textbook data, sourced from scholarly publications, and 69.5\% synthetic data. This synthetic portion is further subdivided into specialized domains, illustrated in Figure \ref{fig:1}, encompassing programming, mathematics, natural language processing, and several other categories that collectively foster both analytical and general knowledge acumen. This structural organization, captured graphically, underpins the dataset's capacity to engender language models with advanced thinking and academic proficiencies, underscoring Intellecta's commitment to developing AI with a profound understanding and versatile reasoning capabilities.

\subsection{Synthetic Data Generation}
The synthetic portion of the dataset is crafted using the advanced Mixtral-8x7B-Instruct-v0.1 model, which is pivotal in generating data that simulates complex thought processes and detailed textbook-like content. This generation phase is designed to fill the gaps in existing datasets by covering a wider array of topics and presenting information in a format that mimics human reasoning and explanation styles. The synthetic data is tailored to enhance the cognitive processing capabilities of AI models, ensuring they can navigate and interpret a multitude of scenarios and subjects effectively.
\subsection{Curation of Intellecta}

The curation of the Intellecta dataset entails a rigorous and systematic process, initiated with the acquisition of raw data from diverse sources. This stage involves OCR \customcite{chaudhuri2017optical} to convert textual content into a structured, machine-readable format, facilitating subsequent computational processes. Simultaneously, raw data for synthetic generation is collected, encompassing a spectrum of domains like coding, language translation, and logical reasoning, which are essential for crafting a multifaceted synthetic dataset using the Mixtral-8x7B-Instruct-v0.1 model.

As a human assimilates information, they not only learn from textbooks but also engage in a complex thought process to understand and apply the knowledge. The Intellecta dataset captures this dual aspect of learning. The synthetic data includes prompts designed to simulate human-like reasoning and textbook-style teaching. Appendix sections \ref{sec:prompt_for_thought} and \ref{sec:prompt_for_textbook}in the paper detail these prompts

During the data refinement stage, the Intellecta dataset is subjected to a thorough cleaning and normalization process, utilizing a customized data-juicer pipeline \customcite{chen2023data} crafted to construct precise data recipes. This phase is pivotal, stripping away extraneous elements such as emails and hyperlinks and standardizing the text to achieve uniformity and coherence throughout the dataset. Employing an array of advanced filters and mappers, the pipeline meticulously addresses factors like alphanumeric balance and the frequency of special characters, tailoring the data to meet rigorous quality standards.

The significance of duplication removal in this context cannot be overstated. Redundant entries not only risk the pre-training phase with instability but also diminish model efficacy due to biased learning outcomes \customcite{kandpal2022deduplicating}. Furthermore, they guard against data leaks that could compromise the validity of benchmark evaluations, particularly for models employing minimal example-based learning. Through the use of Simhash \customcite{sadowski2007simhash}, a method celebrated for its efficiency and fidelity in identifying and purging duplications, the dataset's integrity is fortified, paving the way for more reliable and effective language model training outcomes.

\begin{figure}
\centering
\includegraphics[width=1\textwidth]{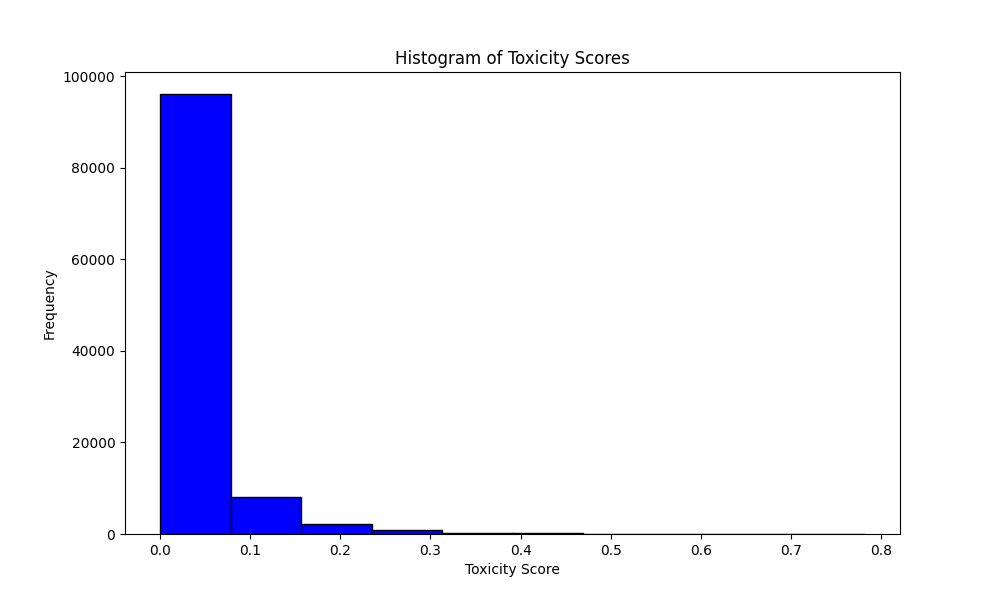}
    \caption{ Histogram of Toxicity Scores:
This histogram represents the frequency distribution of toxicity scores across the dataset, post-screening with the Perspective API. A higher concentration of low toxicity scores illustrates the efficacy of the filtering process in curating an ethically aligned dataset.}
    \label{fig:2}
\end{figure}

The curation process also prioritizes the identification and elimination of toxic content to uphold the ethical standards of the dataset. Utilizing the Perspective API \customcite{hosseini2017deceiving}, the dataset is screened for toxicity, ensuring that the content is free from harmful biases or offensive material \customcite{longpre2023pretrainer}. This step is imperative to ensure the ethical use and application of the dataset in training language models, aligning with the broader objectives of responsible AI development.The result of this process is visually summarized in Figure \ref{fig:2}, showcasing the histogram of toxicity scores across the dataset, where a negligible frequency of high toxicity scores reflects the stringent filtering criteria applied

Diversity in the dataset is visualised through clustering methods, particularly employing DBSCAN \customcite{schubert2017dbscan}to organize data based on semantic similarities. This approach enables the effective grouping of data into thematically consistent clusters, enhancing the dataset's utility for training language models capable of nuanced understanding and generation across varied topics.

The technical orchestration of the Intellecta curation process, from data collection and processing to ethical screening and clustering, reflects a comprehensive and scientifically driven approach. It is designed to yield a dataset that is not only extensive and diverse but also refined to meet the high standards required for advanced language model training. This process underscores the commitment to developing a dataset that serves as a valuable asset for the AI research community, fostering advancements in language modeling and cognitive processing capabilities.

\section{Dataset Description}

\begin{figure}
\centering
\includegraphics[width=1\textwidth]{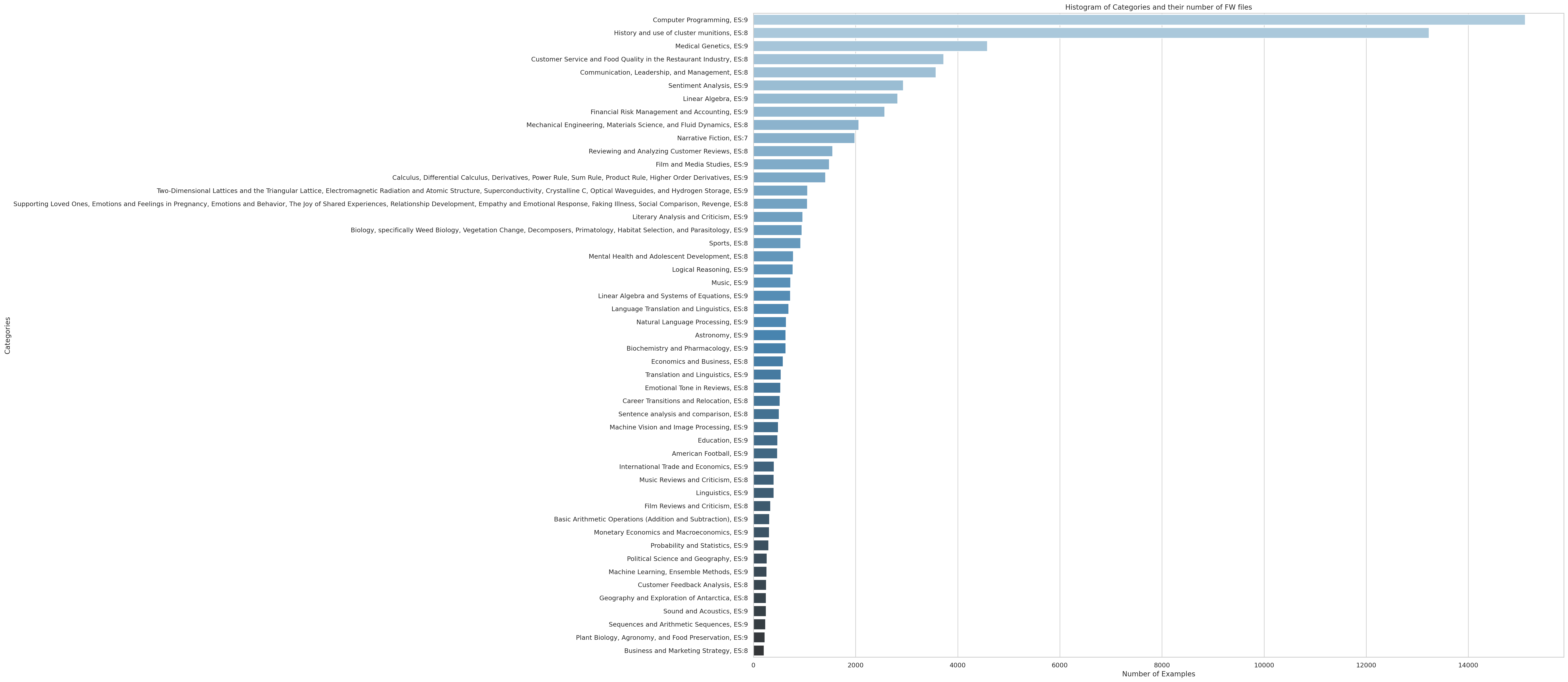}
    \caption{Histogram of Topics in the Intellecta Dataset: This histogram displays the variety of subjects covered, each rated for educational value, highlighting the dataset's commitment to providing a comprehensive learning experience.}
    \label{fig:3}
\end{figure}

The Intellecta dataset is a testament to diversity in machine learning, presenting an extensive range of topics each selected for their substantial educational value (Figure \ref{fig:3}). The selection spans a variety of domains, curated to deepen the pool of training material available for language models.

Prominent among the dataset are topics that hold a high educational value rating of 9, such as Sentiment Analysis, Linear Algebra, and Financial Analysis, reflecting their depth and significance within the compilation. These subjects, together with Narrative Fiction, GUI Programming, and Calculus, form the backbone of the dataset, offering a rich tapestry of academic and practical content. The balance struck across scientific, technical, and literary spheres ensures a rounded representation of knowledge areas.

\begin{figure}
\centering
\includegraphics[width=1\textwidth]{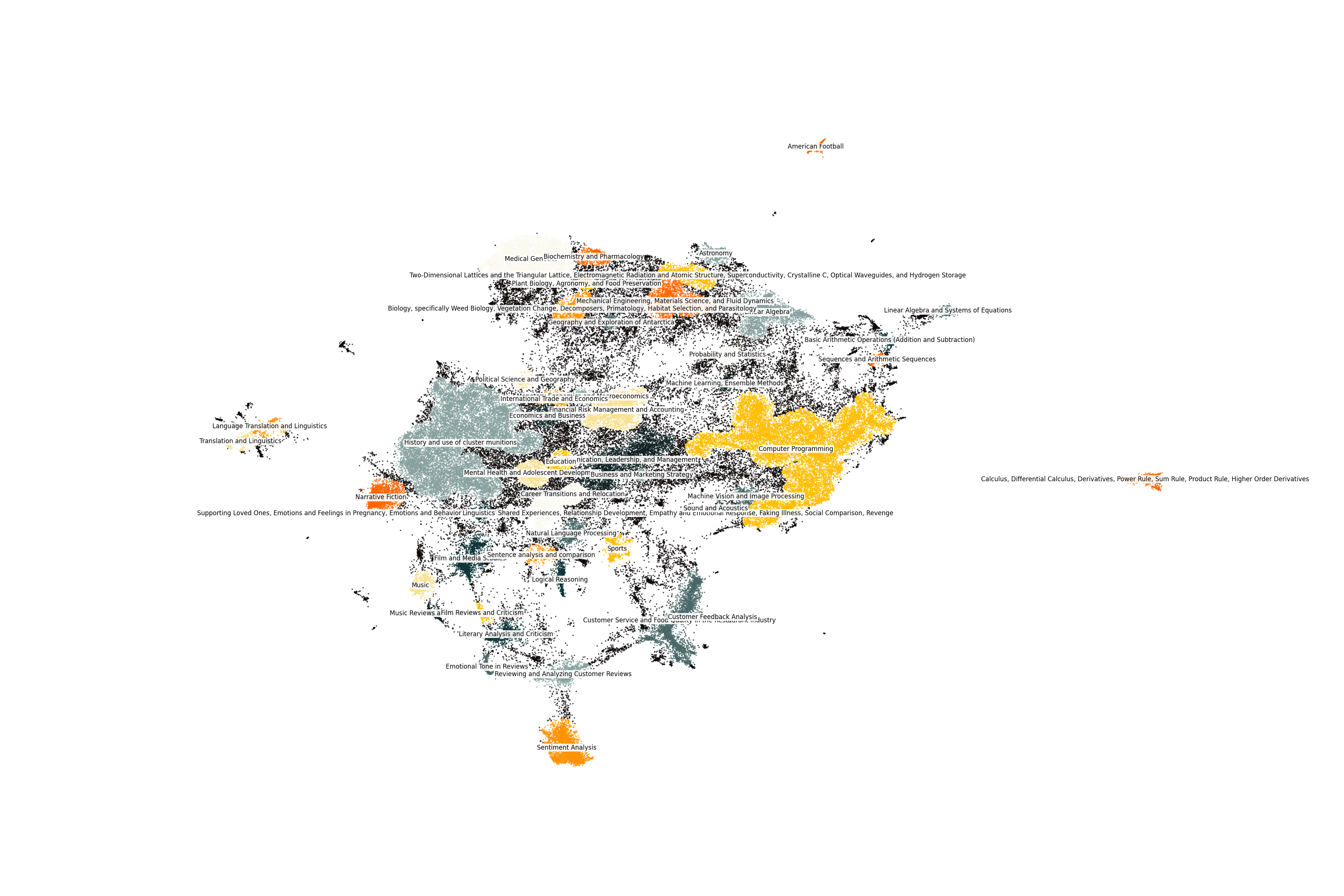}
    \caption{Cluster Analysis of Topics in the Intellecta Dataset: This scatter plot visualizes the semantic clustering of data topics, emphasizing the dataset's depth and the multifaceted nature of the educational content provided.}
    \label{fig:4}
\end{figure}

Venturing into specialized territories, the dataset embraces subjects like Philosophy, Product Reviews, and Food Evaluation, adopting a data curation approach that transcends traditional academic boundaries. The dataset's richness is depicted in Figure \ref{fig:4}, which visualizes the clustering of topics based on diversity and density, highlighting the extensive range of subjects encompassed within the dataset. This visualization captures the dataset's expansive nature, from Ancient History and Project Management to Film Studies and Mathematics, underpinning the potential for language models to apply their capabilities across both technical disciplines and creative endeavors.

Furthermore, practical fields such as Computer Networking, Leadership, and various scientific specializations underscore the dataset's synchrony with the demands of contemporary industry and research landscapes, equipping language models for tangible applications in the real world.

Altogether, Intellecta comprises over 100 individual topics, each chosen and evaluated for their contribution to the educational ecosystem, ensuring that language models trained with this data exhibit comprehensive understanding, robust reasoning, and a spectrum of analytical skills. The strategic amalgamation of these topics not only bolsters the training methodology but also amplifies the potential sophistication of language model outputs, cementing Intellecta's role as an indispensable resource in AI and machine learning research.

\section{Evaluation and Results}

To ascertain the efficacy of the Intellecta  dataset in enhancing language model quality, we embarked on training a 634 million parameter model, referred to as the boomer model, within the bud ecosystem. The model, post-tokenization, processed approximately 11.5 billion tokens, indicative of the vast and comprehensive nature of the data it was trained on. The configuration of the model was meticulously set to optimize its learning and performance capabilities, featuring attributes like a hidden size of 1024, 48 hidden layers, and 16 attention heads, among others, configured within the LlamaForCausalLM architecture.

The evaluation of the Intellecta  model's performance(Table \ref{tab:AppendixTable}) was comprehensive, involving benchmarking across a variety of tasks to assess its generalization ability across domains. This was crucial in determining the model's proficiency in handling diverse data types and its adaptability to various problem-solving scenarios.

\begin{table}[ht]
\centering
\scalebox{0.8}{ 
\footnotesize 
\begin{tabular}{|p{5cm}|p{2cm}|p{1.5cm}|p{1.5cm}|p{2cm}|p{1.5cm}|p{2cm}|p{1.5cm}|}

\hline
\textbf{Model} & \textbf{Parameters} & \textbf{Token} & \textbf{ARC} & \textbf{HellaSwag} & \textbf{MMLU} & \textbf{Winogrande} & \textbf{GSM8K} \\
\hline
EleutherAI/pythia-1b-deduped & 1.1B & - & 29.10 & 49.65 & 24.27 & 53.59 & 1.14 \\
facebook/opt-1.3b & 1.3B & 180B & 29.52 & 54.53 & 24.96 & 59.75 & 0.15 \\
Qwen/Qwen1.5-0.5B & 620M & - & 31.48 & 49.05 & 39.35 & 57.22 & 16.3 \\
HuggingFaceTB/cosmo-1b & 1.8B & 30B & 38.57 & 55.13 & 26.69 & 55.49 & 5.53 \\
TinyLlama/TinyLlama-1.1B-Chat-v0.6 & 1.1B & 3T & 31.66 & 55.79 & 25.98 & 59.35 & 2.12 \\
\textbf{boomer-634m} & \textbf{634M} & \textbf{11.5B} & \textbf{29.86} & \textbf{39.24} & \textbf{25.91} & \textbf{50.61} & \textbf{1.67}  \\
EleutherAI/gpt-neo-1.3B & 1.3B & 380B & 31.23 & 48.47 & 24.82 & 56.91 & 0.45 \\
\hline
\end{tabular}}
\caption{Model performance on various benchmarks}
\label{tab:AppendixTable}
\end{table}

The boomer model, underwent rigorous testing against several benchmarks, yielding the following results:

For comparative analysis, we examined similar parameter models trained on different large datasets. Notable among these are models like EleutherAI/pythia-1b-deduped, facebook/opt-1.3b, and HuggingFaceTB/cosmo-1b, each demonstrating varying levels of performance across the benchmarks.

Our analysis highlighted that the boomer-634m model, despite having only 634 million parameters and being trained on 11.5 billion tokens, exhibited competitive performance across multiple tasks. For instance, in the ARC and HellaSwag benchmarks, it showcased comparable, if not superior, outcomes against models with larger datasets and parameters. This indicates the Intellecta  dataset's potential to produce high-quality language models capable of significant cross-domain generalization, even with comparatively fewer parameters and token count.

The evaluation of the Intellecta  model underscores the dataset's capacity to facilitate the training of potent language models, achieving commendable performance across diverse benchmarks. The results affirm the boomer dataset's role in driving forward the capabilities of language models, reinforcing its value in the field of AI and language processing research.

\section{Conclusion}
The Intellecta  dataset signifies a transformative stride in synthetic data generation, engineered to enhance the cognitive capabilities of large language models (LLMs). Through its strategic design and rigorous curation process, boomer not only aims to replicate the success of high-performing LLMs but also to surpass it, establishing a new benchmark in the realm of AI research. The dataset’s diversity, scale, and commitment to quality and ethical standards have positioned it as a critical resource capable of driving substantial advancements in machine learning and AI.

The empirical evaluation of the Intellecta  model underscores the dataset’s effectiveness, showcasing its potential to produce robust language models with commendable cross-domain generalization. Despite the constraints of data volume and model size, the boomer dataset demonstrated competitive performance across varied benchmarks, reflecting its high-quality, nuanced, and contextually rich content. This achievement highlights the efficacy of our curation methodology, which, when expanded, promises to yield even greater results in language model training.

Looking ahead, the scalable and inclusive design of the Intellecta  dataset offers ample opportunity for growth and enhancement. By persisting in our meticulous approach to data curation and quality assurance, we anticipate generating a more voluminous and enriched dataset that could further elevate the performance of language models. Thus, boomer stands not just as a product of current research endeavors but as a beacon for future innovations in the AI landscape, promising to unlock unprecedented capabilities in language modeling and cognitive processing.

\section{References}
\bibliographystyle{plainnat} 
\bibliography{tempbib} 

\section{Appendix}
\subsection{Prompt for Thought data}

Here the model is guided to emulate an expert thinker and philosopher's cognitive process, breaking down instructions into logical steps and reasoning chains. It generates not only the steps towards an answer but also rules and exceptions learned from the problem-solving journey
\label{sec:prompt_for_thought}
\begin{mdframed}[backgroundcolor=promptbackground, linecolor=promptframe, leftmargin=10, rightmargin=10, innerleftmargin=10, innerrightmargin=10, innertopmargin=10, innerbottommargin=10, linewidth=1pt]
\textbf{Prompt:}
\begin{lstlisting}
You are an expert thinker, problem solver,  philosopher, generalist, Logician, coder, mathematician, and scientist with extreme attention to details. You are tasked with creating reasoning steps or thought processes, applicable rules, do not do rules list (Things that must not be done), and generalizable rules (Rules learned from this problem that can be applied to other problems and situations). Following are the steps that you can take to successfully finish the task. First, consider the instruction (Instruction) given and the response (Response) provided carefully to fully understand the problem & solution, Take a deep breathe and Let us think step by step to make sure that we arrive at the answer from the provided instruction, without any changes to the final answer.
Following are the steps to be taken in order to create a detailed thought process and reasoning steps - First, carefully analyze the instruction & the given human response to extract the key information components and break it down into logical sub-questions. This helps set up the framework for reasoning. The goal is to construct an internal search tree. - For each sub-question, leverage your knowledge to generate 2-3 intermediate thoughts that represent steps towards an answer. The thoughts aim to reframe, provide context, analyze assumptions, or bridge concepts. - Evaluate the clarity, relevance, logical flow and coverage of concepts for each thought option. Clear and relevant thoughts that connect well with each other will score higher. - Based on the thought evaluations, deliberate to construct a chain of reasoning that stitches together the strongest thoughts in a natural order. - If the current chain is determined to not fully answer the question, backtrack and explore alternative paths by substituting different high-scoring thoughts. - Throughout the reasoning process, aim to provide explanatory details on the thought process rather than just state conclusions, including briefly noting why some thoughts were deemed less ideal. - Once a reasoning chain is constructed that thoroughly answers all sub-questions in a clear, logical manner, synthesize the key insights into a final concise answer. - Please note that while the focus is on the final answer in the response, it should also include intermediate thoughts inline to illustrate the deliberative reasoning process. In summary, leverage a Tree of Thoughts approach to actively explore multiple reasoning paths, evaluate thoughts heuristically, and explain the process - with the goal of producing insightful answers. After figuring out the final reasoning that helps with the problem solving, write down a list of Rules learned from the problem solving that allows for solving similar problems in the future, also mention a list of what not to do as rules - a list of things that should not be done on scenarios like these to avoid any errors or undesirable outputs.

Instruction: {instruction}
\end{lstlisting}
\end{mdframed}

\subsection{Prompt for Textbook data}

It directs the model to function as an educator, dissecting a problem to unearth and elaborate on underlying concepts in a structured, pedagogical manner. The model creates chapters with clear, concise explanations, examples, and exercises, starting from basic to complex ideas, replicating the educational narrative of a textbook.
\label{sec:prompt_for_textbook}

\begin{mdframed}[backgroundcolor=promptbackground, linecolor=promptframe, leftmargin=10, rightmargin=10, innerleftmargin=10, innerrightmargin=10, innertopmargin=10, innerbottommargin=10, linewidth=1pt]
\textbf{Prompt:}
\begin{lstlisting}

You are an expert teacher, book writer, mathematician, logician etc. Given below is the instruction (Instruction) provided by a user and the response(Response) to that instruction provided by another human. Take a look at the problem and the response to identify the key concepts required to solve the problem. Once the key concepts, ideas, facts, theories etc relevant to the problem are identified, take a deep breath and think step by step to decompose them into more fundamental concepts or ideas, these must be written in the format of a textbook with a chapter detailing the particular concept, every chapter must have an easy to understand, clear concise description with relevant examples, use-cases, exercises, applications, etc. The textbook should cover ideas from the most simple concepts to the most complex ones.

Instruction: {instruction} 
\end{lstlisting}
\end{mdframed}

\end{document}